\pdfoutput=1

\documentclass[11pt]{article}

\usepackage[preprint]{acl}

\usepackage{times}
\usepackage{latexsym}

\usepackage[T1]{fontenc}

\usepackage[utf8]{inputenc}

\usepackage{microtype}

\usepackage{inconsolata}

\usepackage[most]{tcolorbox}
\usepackage{graphicx,amssymb}
\usepackage{multirow}
\usepackage{array}
\usepackage{booktabs}
\usepackage{amsfonts}
\usepackage{bbding}
\usepackage{bm}
\usepackage{color,colortbl,xcolor}
\usepackage{stfloats}
\usepackage{enumitem}
\usepackage{nccmath}
\usepackage{subfigure}
\usepackage{soul}
\usepackage{pifont}

\definecolor{hlgreen}{HTML}{B2D5CB}
\definecolor{hlblue}{HTML}{ADD8E6}
\definecolor{bggrey}{HTML}{5E5D65}
\definecolor{bgpink}{HTML}{CEAEB9}
\definecolor{bgblue}{HTML}{8D91AA}

\title{SingaKids: A Multilingual Multimodal Dialogic Tutor \\for Language Learning}

\author{
\textbf{
Zhengyuan Liu\textsuperscript{\ding{71}}
\ \quad
Geyu Lin\textsuperscript{\ding{71}}
\ \quad
Hui Li Tan\textsuperscript{\ding{71}}
\ \quad
Huayun Zhang\textsuperscript{\ding{71}}}\\
\textbf{
\ \quad
Yanfeng Lu\textsuperscript{\ding{71}}
\ \quad
Xiaoxue Gao\textsuperscript{\ding{71}}
\ \quad
Stella Xin Yin\textsuperscript{\ding{118}}}\\
\textbf{
\ \quad
He Sun\textsuperscript{\ding{86}}
\ \quad
Hock Huan Goh\textsuperscript{\ding{86}}
\ \quad
Lung Hsiang Wong\textsuperscript{\ding{86}}
\ \quad
Nancy F. Chen\textsuperscript{\ding{71}}}\\
\textsuperscript{\ding{118}}Nanyang Technological University, Singapore\\
\textsuperscript{\ding{86}}National Institute of Education (NIE), Singapore\\
\textsuperscript{\ding{71}}Institute for Infocomm Research (I$^2$R), A*STAR, Singapore\\
\texttt{\{liu\_zhengyuan,hltan,nfychen\}@i2r.a-star.edu.sg}
}

\begin{document}
\maketitle

\begin{abstract}
The integration of generative artificial intelligence into educational applications has enhanced personalized and interactive learning experiences, and it shows strong potential to promote young learners language acquisition. However, it is still challenging to ensure consistent and robust performance across different languages and cultural contexts, and kids-friendly design requires simplified instructions, engaging interactions, and age-appropriate scaffolding to maintain motivation and optimize learning outcomes.
In this work, we introduce SingaKids, a dialogic tutor designed to facilitate language learning through picture description tasks. Our system integrates dense image captioning, multilingual dialogic interaction, speech understanding, and engaging speech generation to create an immersive learning environment in four languages: English, Mandarin, Malay, and Tamil. We further improve the system through multilingual pre-training, task-specific tuning, and scaffolding optimization. Empirical studies with elementary school students demonstrate that SingaKids provides effective dialogic teaching, benefiting learners at different performance levels.
\end{abstract}

\section{Introduction}
The integration of generative artificial intelligence into educational technologies has significantly transformed learning environments by enabling more personalized and adaptive experiences \cite{zhang2021ai,yan2024practical}. These AI-driven systems can respond to individual learner needs, provide immediate feedback, and create engaging interactions that support knowledge acquisition and skill development \cite{zhai2021review}. In the domain of language learning, this technological advancement presents particularly promising opportunities, especially for young learners who benefit from interactive and contextually rich learning experiences \cite{pokrivvcakova2019preparing, ji2023systematic}.

\begin{figure}[t!]
\centering
\includegraphics[width=0.92\linewidth]{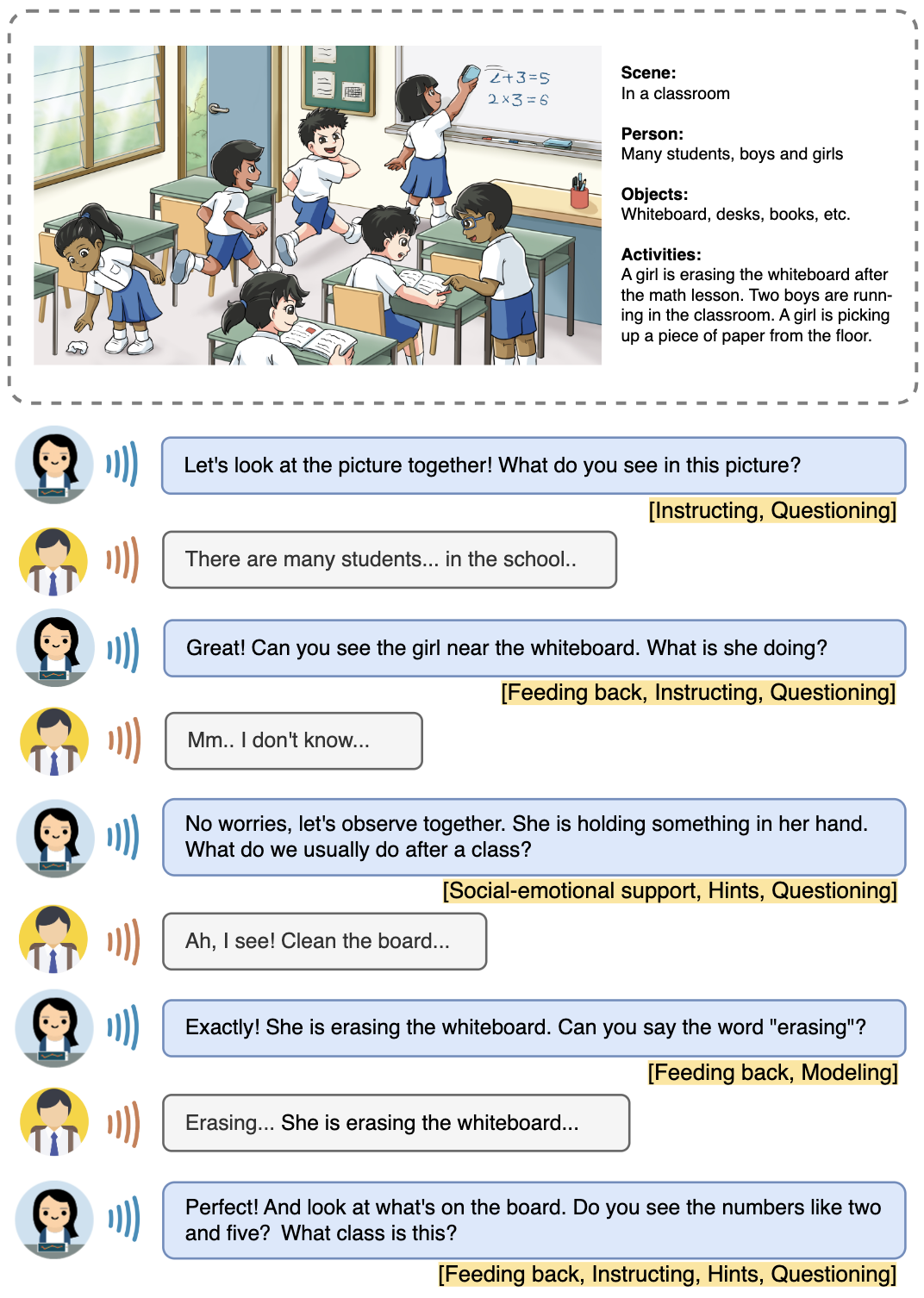}
\caption{Multi-modal dialogic interaction for language learning through the image description task. Students use speech to interact with the system. Pink spans denote the dynamic scaffolding strategies.}
\label{fig:intro_example}
\vspace{-0.3cm}
\end{figure}

Recent advances in large language models (LLMs) and multimodal systems have demonstrated impressive capabilities in understanding and generating human language across diverse contexts \cite{achiam2023gpt4,team2023gemini}. However, deploying these technologies effectively for educational purposes, particularly for children's language acquisition, presents several significant challenges. First, ensuring consistent performance across different languages and cultural contexts remains difficult, as most systems exhibit stronger capabilities in high-resource languages like English compared to others \cite{wang2023seaeval}. Second, designing child-friendly interactions requires careful consideration of cognitive load, attention spans, and developmental appropriateness—factors that often necessitate simplified instructions, engaging dialogue patterns, and age-appropriate scaffolding to maintain motivation and optimize learning outcomes \cite{liu2024scaffolding}.

To address these challenges, we introduce SingaKids, a dialogic tutor specifically designed to facilitate language learning through picture description tasks. The oral practice enhances children's language acquisition by stimulating vocabulary development, syntactic complexity, and  observational skills, and facilitating contextual language use within meaningful visual contexts - all essential components of early linguistic competence development. To this end, our system integrates four components: (1) dense image captioning to provide rich visual context understanding, (2) multilingual dialogic interaction to support natural conversational flow, and deliver appropriate feedback and guidance, (3) robust speech understanding to process young learners' verbal responses, and (4) kids-friendly speech generation to improve the student engagement during tutorial sessions. SingaKids operates across four languages relevant to Singapore's multicultural context: English, Mandarin, Malay, and Tamil, making it accessible to students from diverse linguistic backgrounds.

We further enhanced the system's performance through multilingual pre-training strategies, task-specific tuning to optimize picture description dialogue flows, and scaffolding optimization to provide appropriate levels of support based on learner responses. This approach allows the system to adapt its interaction patterns to match learners' proficiency levels and specific linguistic needs.
To evaluate the effectiveness of SingaKids, we conducted empirical studies with first and second-grade elementary school students of different language proficiency levels. Our findings demonstrate that the system provides effective dialogic teaching experiences that support language acquisition through natural conversation about visual stimuli. Notably, students at various performance levels showed improvements in descriptive language skills, vocabulary usage, and conversational fluency after engaging with the system.

This work contributes to the growing field of AI-enhanced language education by demonstrating how multimodal, multilingual systems can be successfully deployed to support young learners' language development. By addressing the challenges of cross-linguistic consistency and age-appropriate interaction design, SingaKids represents a step forward in creating accessible and effective learning agents for diverse educational contexts.

\begin{figure*}[t!]
\centering
\includegraphics[width=1.0\linewidth]{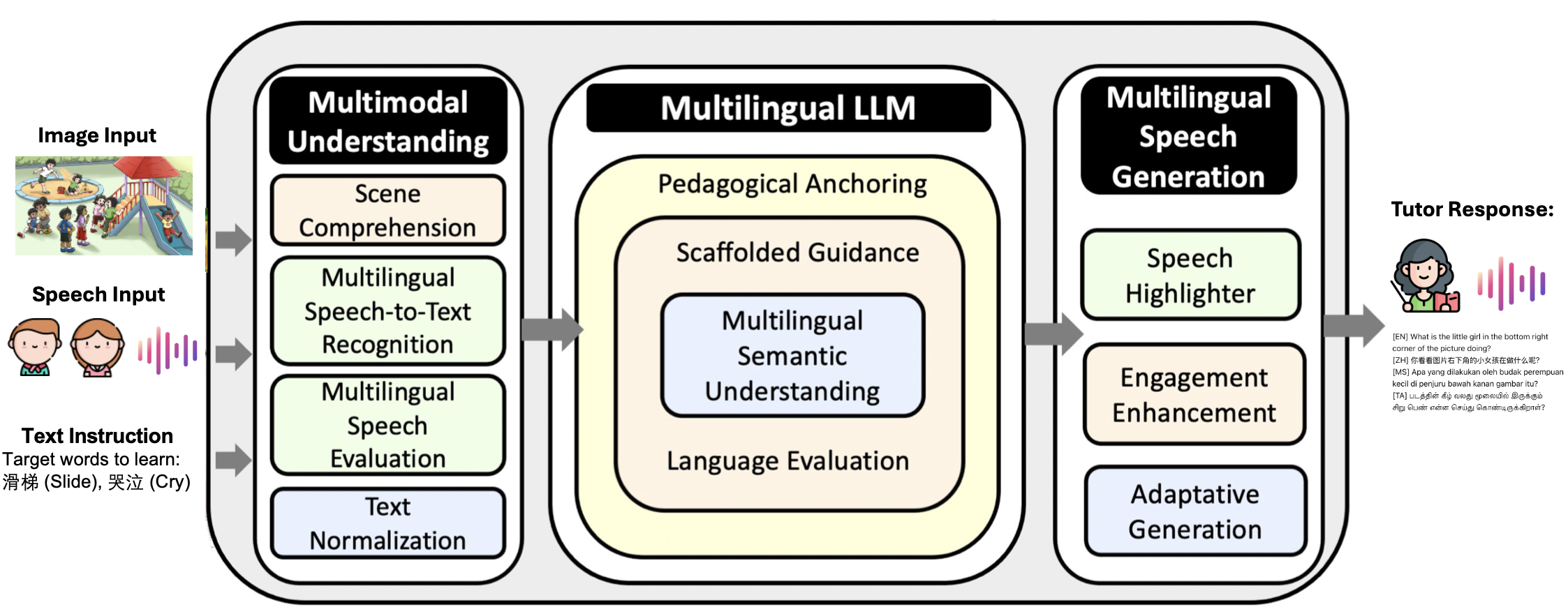}
\caption{Overview of the conversational tutor architecture for language learning via the image description task.}
\label{fig:framework}
\vspace{-0.1cm}
\end{figure*}

\section{Related Work}
Intelligent tutoring systems aim to replicate human tutoring by providing personalized instruction and adaptive feedback to language learners. The advancement of ITSs has marked a significant step forward in education practice \cite{Graesser2018,demszky2023ncte,wang2023step}. These systems provide personalized learning experiences and instant feedback \cite{Chaffar2004,Harley2015,Grivokostopoulou2017}, tailored to learners' characteristics and needs \cite{Dzikovska2014,Grawemeyer2016,Nihad_El2017}, and are shown to positively influence students' engagement in learning and academic performance \cite{Kulik2016,Xu2019}.

Dialogue tutor is a particular type of intelligent tutoring system that interacts with students via natural language conversation \cite{nye2014autotutor,ruan2019quizbot}. In STEM domains, conversational ITSs can facilitate university students in problem-solving by providing real-time feedback and hints in text formats \cite{nye2023generative,paladines2020systematic, arnau2023methodological}.
However, prior work has widely relied on rule-based systems with human-crafted domain knowledge \cite{nye2014autotutor,Graesser2018}, or data-driven approaches that require a certain amount of human annotation for supervised learning \cite{maclellan2022domain}. Recently, LLMs show strong potential to build dialogue tutors with less data supervision and higher coherence \cite{afzal-etal-2019-development,demszky2023ncte,macina2023opportunities}, and they can be further improved by integrating LLMs with pedagogical and learning science principles \cite{stasaski2020cima,sonkar2023class,macina2023mathdial}.

\section{SingaKids System Architecture}
In a picture description session, teachers first present an image and ask students to observe it carefully. They pose open-ended questions like ``\textit{What do you see in this picture?}'' to stimulate observation, then guide students beyond basic object identification to describe qualities using adjectives and adverbs, enhancing vocabulary, organization, and fluency. The activity concludes with introducing new vocabulary and encouraging students to create stories about the image, developing creativity and narrative skills.

Drawing from real-world teaching sessions, the overall architecture of our conversational tutoring system is illustrated in Figure \ref{fig:framework}. Interactions begin with the Multimodal Understanding:
the scene comprehension extracts the keywords, objects, and events in the given picture \cite{densecap};
the multilingual speech recognition converts the student's spoken response into text;
speech evaluation component is to assess the student's oral language proficiency \cite{wong2022variations}.

The Multilingual LLM represents the core of educational interaction:
multilingual semantic understanding interprets the student's response in context;
language evaluation assesses the linguistic accuracy and completeness of their description;
scaffolded guidance determines the appropriate level of support needed; This component effectively analyzes the student's current understanding and formulates an appropriate teaching strategy;
pedagogical anchoring establishes high-level educational objectives such as word understanding or sentence construction.
Moreover, for elementary grade 1 and 2, we evaluate students' skills in making sentences to describe the activities in the image, focusing on their vocabulary usage. The language evaluation can be adapted for higher grade levels, by measuring grammatical correctness and coherent narratives \cite{genishi2015children}.

The system's response is formed in both text and audio outputs. The Multilingual Speech Generation converts text utterance into natural and engaging speech to maintain student motivation \cite{kim2021conditional};
In addition, beyond simple text-to-speech synthesis, we incorporate a highlight component which can emphasize important keywords or pronunciation errors \cite{zhang2021multilingual}.

Throughout this pipeline, the system maintains an educational dialogic flow, asking guiding questions, providing hints, offering corrections, and acknowledging progress as needed. If a student struggles with specific vocabulary when describing the image (for example, using general terms like \textit{``playing''} instead of specific verbs like \textit{``swimming''} or \textit{``climbing''}), the system will scaffold their learning through targeted questions and gradually decreasing support until they can independently produce the target words \cite{liu2024scaffolding}.

\section{Module Optimization}
Young students at the early elementary stage with limited language proficiency raise unique requirements for human-AI interaction. Enhancing the multilingual capability of the core components can improve communication efficacy as well as handling mixed language usage or intra-sentential code-switching. This is particularly important in environments where learners may express themselves in multiple languages they are exposed to at home or school. Additionally, scaffolding kids requires simplified instructions, and consistent engagement through positive feedback and social-emotional support. While maintaining reasonable performance in English and Mandarin, we specifically focus on improving Malay and Tamil to better serve Singapore's multilingual student population.\footnote{We used the Huggingface codebase for model training \& evaluation (https://github.com/huggingface/transformers). All experiments were conducted on Nvidia A100 40/80GB GPUs.} 

\begin{figure}[t!]
\centering
\includegraphics[width=0.97\linewidth]{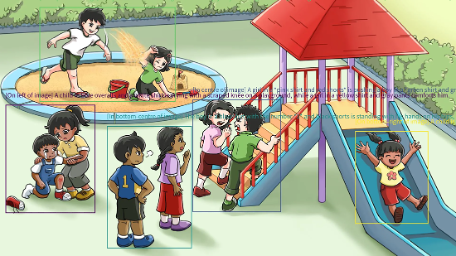}
\includegraphics[width=1.0\linewidth]{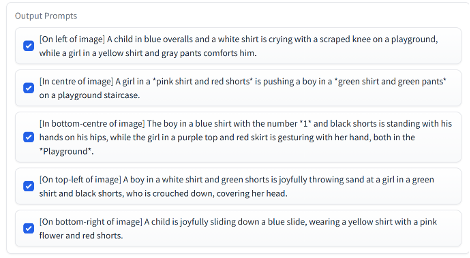}
\caption{Dense image captioning with contextualization (caption of each event aligns with large narrative) and fine-grained understanding (detailed description of objects, characters and activities).}
\label{fig:vision_model}
\vspace{-0.2cm}
\end{figure}

\subsection{Fine-grained Image Description}
For the picture-guided conversation flow, we propose a dense image captioning module for visual storytelling. The goals are to identify the key events of interest in the image as well as generate rich captions for each key event of interest. Referring to the example in Figure \ref{fig:vision_model}, the caption for each event shall be aligned with the larger narrative of the image (better contextualization), and include detailed description of the objects, characters, and activities (fine-grained understanding). State-of-the-art multi-modal LLMs (MLLMs), especially smaller size models, generally struggle with dense image content. The MLLMs often generate general and broad descriptions of the image, and are limited in deeper analysis of the visual details. Moreover, hallucinations occur due to complex or ambiguous image content. Hence, we adopt a two-stage approach – event bounding box proposal and caption generation. For event bounding box proposal, we leverage  robust person and object detection \cite{liu2024grounding}, human segmentation \cite{kirillov2023segment}, coupled with depth estimation \cite{bhat2023zoedepth}, for probabilistic reasoning on the neural detections. For caption generation, we use chain-of-thought prompting on a MLLM, InternVL2.5 \cite{chen2024internvl}, to incorporate global context understanding into the individual event captions. We achieved a 75\% sentence-level accuracy in our image testbed, which provides reasonable content for the conversational process.

\begin{figure}[t!]
\centering
\includegraphics[width=0.89\linewidth]{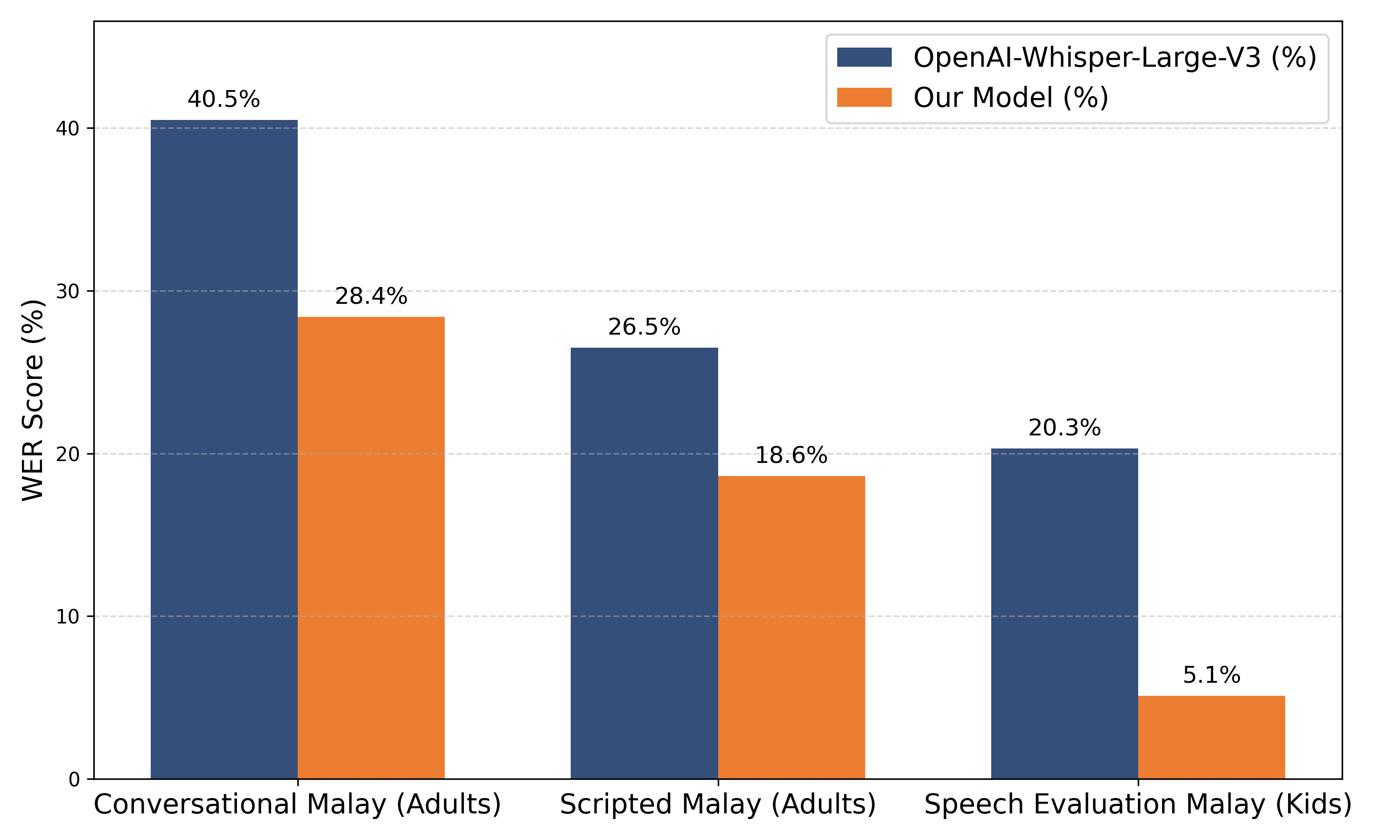}
\caption{Malay ASR evaluation results.}
\label{fig:asr_malay}
\vspace{-0.3cm}
\end{figure}

\begin{figure}[t!]
\centering
\includegraphics[width=0.89\linewidth]{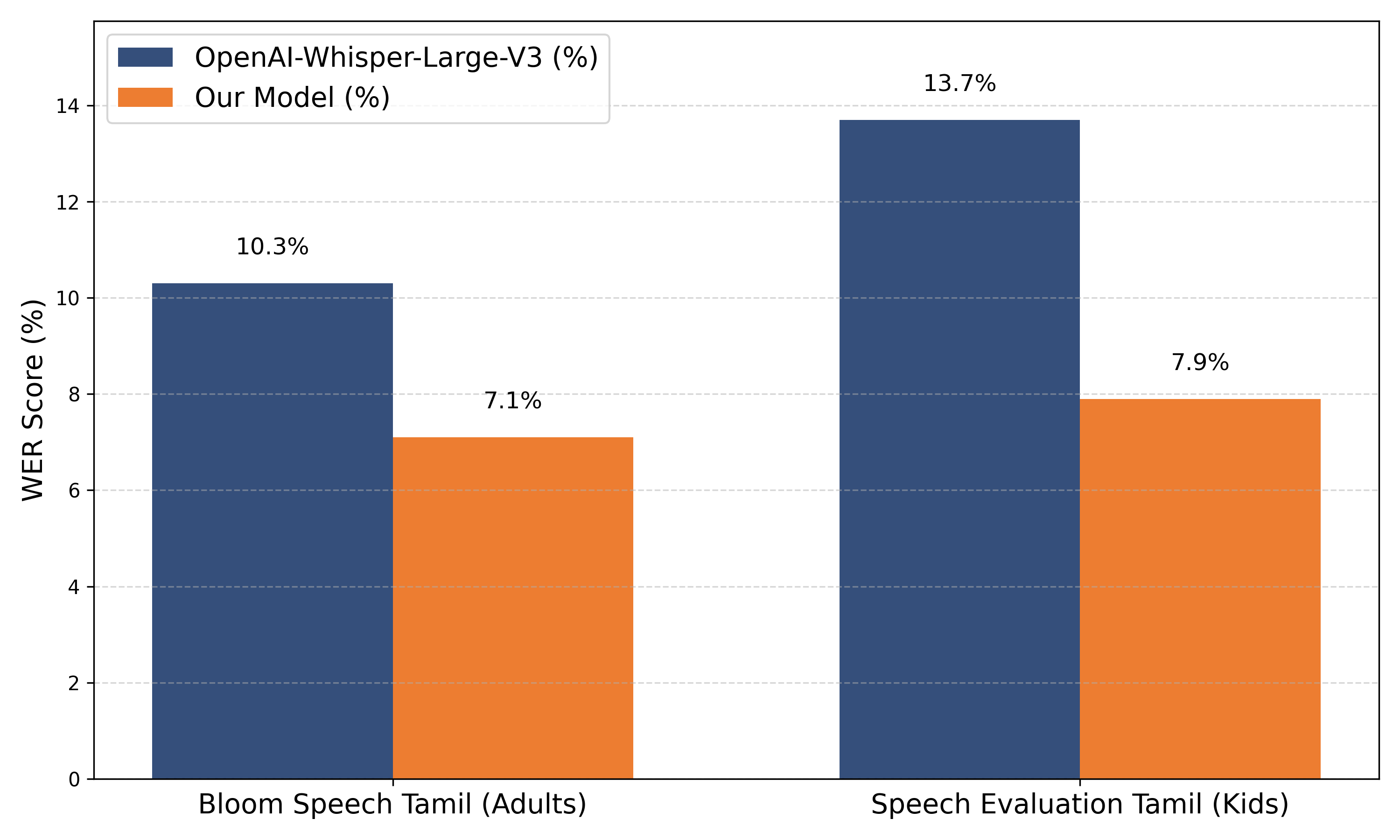}
\caption{Tamil ASR evaluation results.}
\label{fig:asr_tamil}
\vspace{-0.3cm}
\end{figure}

\subsection{Improving Multilingual ASR}
To enhance the multilingual ASR capabilities of our system, we selected Whisper-large-V3 \cite{radford2022whisper} as the base model and fine-tuned it with additional Malay and Tamil speech data. Preliminary analysis revealed significant performance gaps when processing lower-resource languages (e.g., Malay, Tamil), and in children’s voice transcribing \cite{attia2024kid}. We addressed this limitation by gathering a local dataset comprising 2,800-hour Tamil recordings and 1,000-hour Malay recordings from more than 1,000 native speakers from different age groups and linguistic contexts.

As shown in Figure \ref{fig:asr_malay} and Figure \ref{fig:asr_tamil}, we compare Whisper-large-V3 with our fine-tuned model on Malay and Tamil data. Evaluating on Malay data, we achieved a lower WER from 40.5\% to 28.4\% on conversational speech and from 20.3\% to 5.1\% on children speech \cite{zhang2021multilingual}. For Tamil, we achieved lowers WER from 10.3\% to 7.1\% on Bloom Speech Tamil \cite{leong2022bloom} and from 13.7\% to 7.9\% on a children speech data \cite{zhang2021multilingual}. We obtained consistent improvements at all test sets, particularly in children's voice.

\begin{figure}[t!]
\centering
\includegraphics[width=1.0\linewidth]{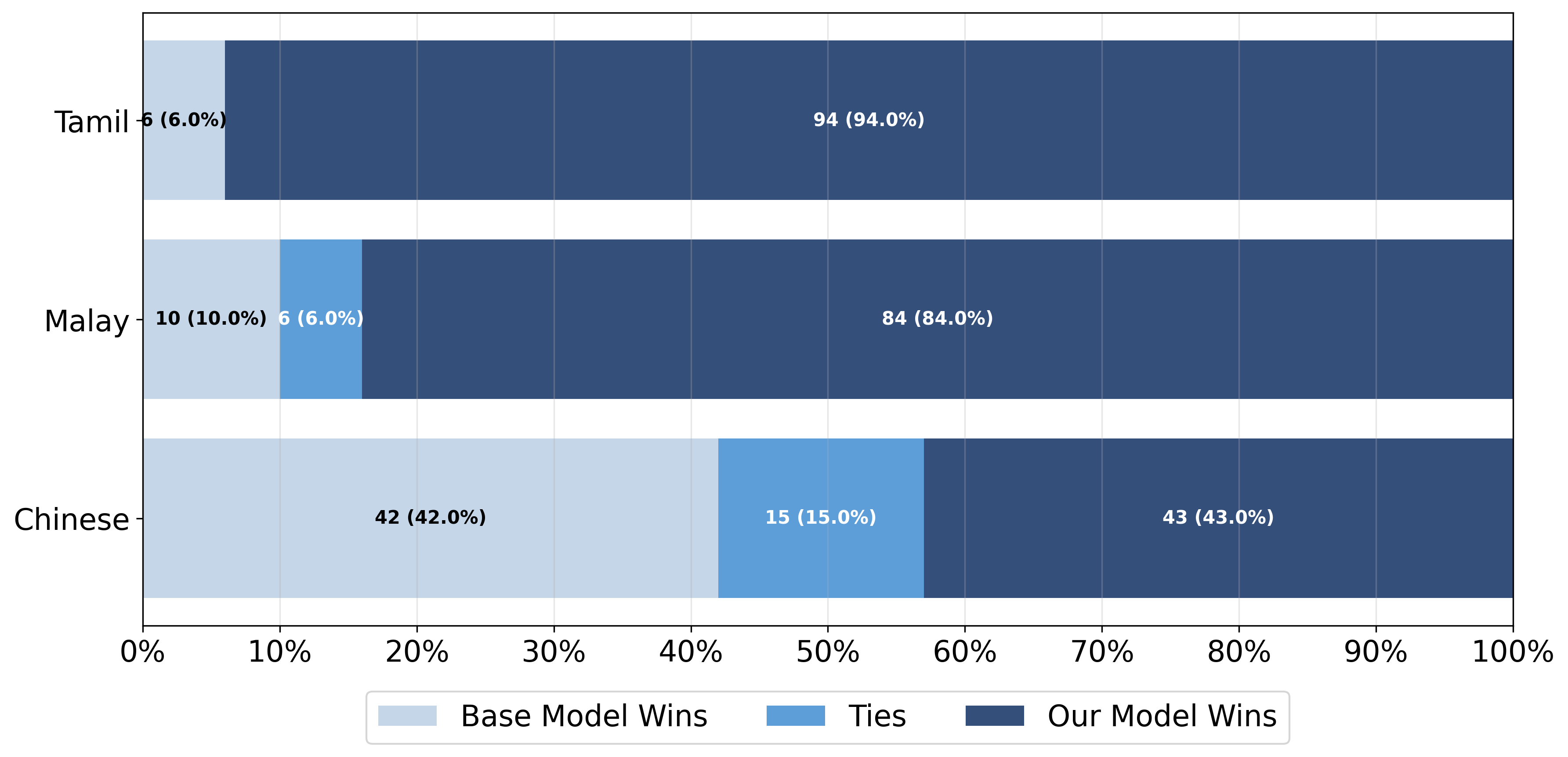}
\caption{Comparison between the base model and our model of translation capability.}
\label{fig:elo_translation}
\vspace{-0.1cm}
\end{figure}

\begin{figure}[t!]
\centering
\includegraphics[width=1.0\linewidth]{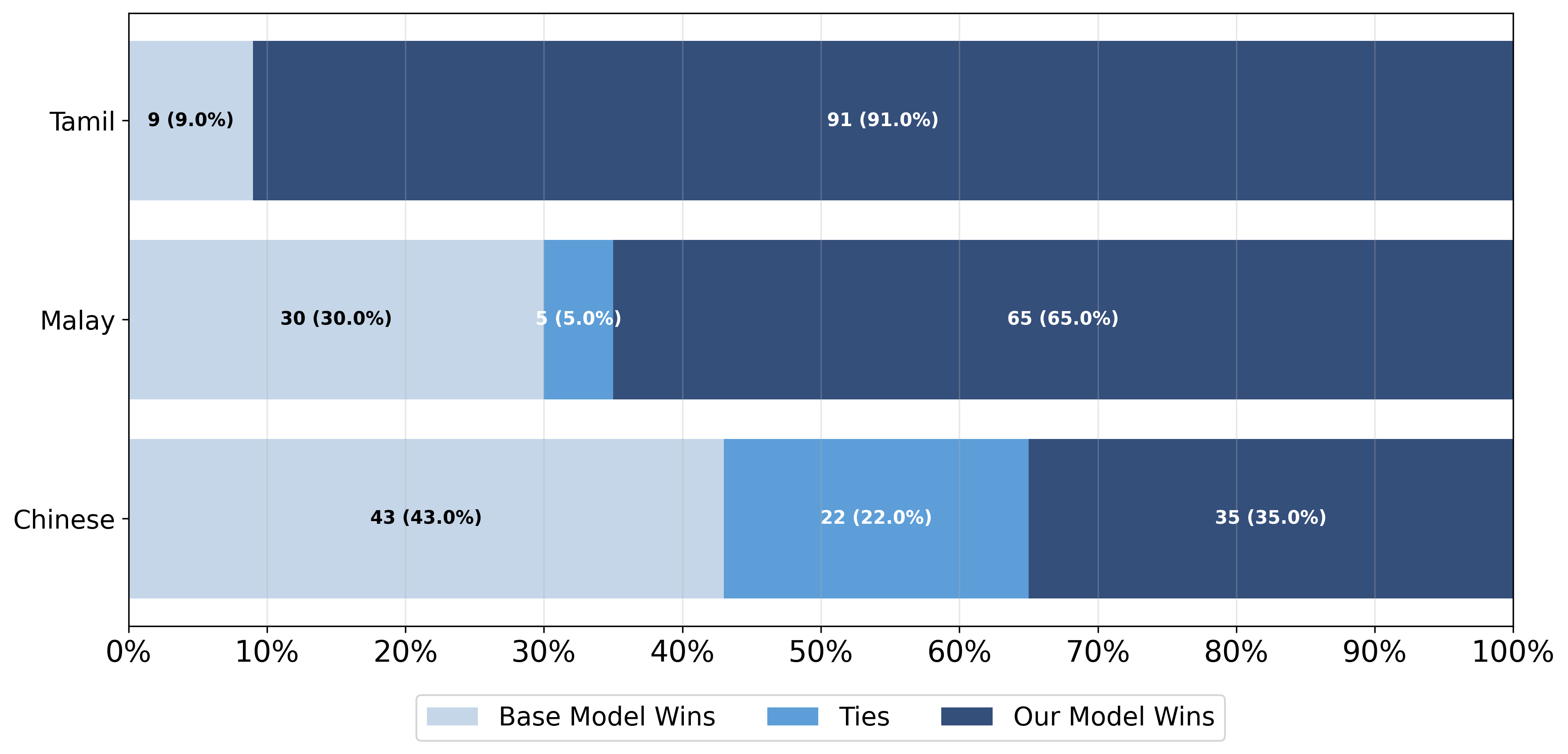}
\caption{Comparison between the base model and our model of multilingual instruction following.}
\label{fig:elo_instruction}
\end{figure}

\subsection{Improving Dialogue LLM}
We improve the dialogue component built on a text LLM from the following two aspects:

\subsubsection{Multilingual Capability}
LLMs often show downgraded performance in low-resource languages, and this problem becomes more prominent on smaller models. In this work, we selected Qwen1.5-4B \cite{qwen} as the base model for a balance of performance and cost-efficiency.\footnote{We tested a set of Qwen1.5 models from 1.8B to 14B, and observed that model size is strongly correlated with multilingual capabilities, especially for languages with lower resources such as Malay and Tamil.}
Our multilingual optimization follows a two-stage process:
First, we conducted continue pre-training on 14B tokens of mixed data with four languages (English, Mandarin, Malay, Tamil) \cite{penedo2024fineweb}. We set a balanced sampling rate to elevate the multilingual modeling of Malay and Tamil, and English and Mandarin data play a role to retain the fundamental language capabilities. Second, we enhanced the model's multilingual instruction following by multi-task learning \cite{OpenHermes-2.5} and cross-lingual alignment \cite{muennighoff2023crosslingual,lin2025crossin}, including multilingual role-play corpora generated through simulating diverse conversation scenarios \cite{sun2024parrot,liu-etal-2024-optimizing}.
To further strengthen cross-lingual capabilities, we did a hybrid training approach that combines translation and cross-lingual problem-solving tasks \cite{muennighoff2022crosslingual,liu2022singlish}. This enables the language model of better semantic fusion across languages. Experimental results shown in Figure \ref{fig:elo_translation} and Figure \ref{fig:elo_instruction} show improvement on multilingual translation and instruction following.

\subsubsection{Scaffolding-guided Augmentation}
\begin{figure}[t!]
\centering
\includegraphics[width=1.0\linewidth]{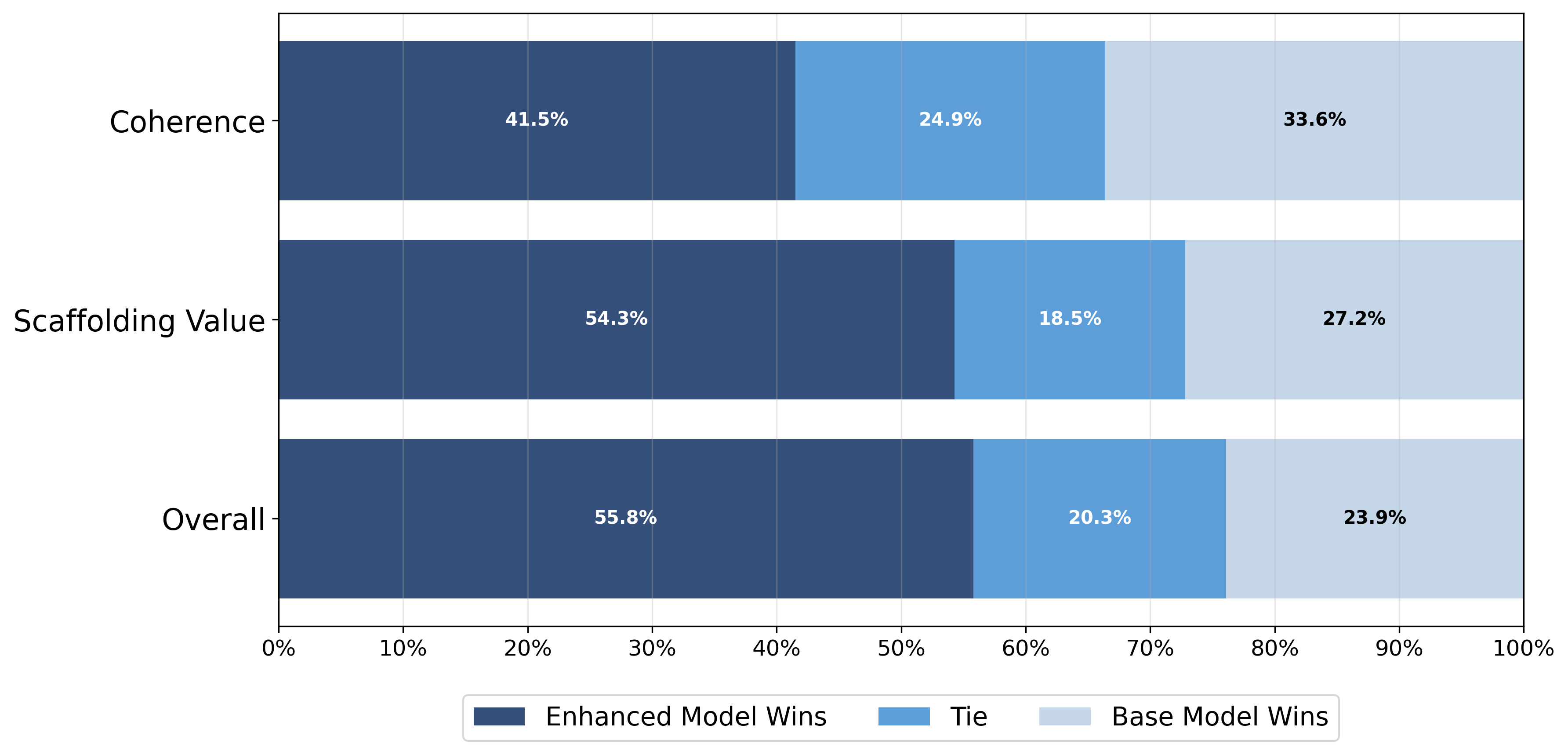}
\caption{Comparison between the base dialogue model and pedagogical-enhanced model through the LLM-as-a-judge evaluation.}
\label{fig:elo_dialogue}
\end{figure}

We improved the dialogue model's pedagogical effectiveness by training with scaffolding instructions and personality-aware student simulation \cite{de2023interactional,sonkar2023class,liu2024scaffolding,liu-etal-2024-personality}. Our scaffolding framework is formulated upon the dialogic teaching theory \cite{Alexander2006}, where the tutor encourages exchange of ideas using follow-up questions, clues, elaborations, or recaps. We conducted a theory-inspired practice by sampling synthetic dialogue samples from a stronger teacher LLM (\texttt{GPT-4} \cite{achiam2023gpt}) to guide the smaller LLMs, which is capable of providing scaffolded interactions based on learners response.
Moreover, in dialogic teaching, recognizing and adapting to individual characteristics can significantly enhance student engagement and learning efficiency. We built a taxonomy of student personality profiles based on established traditional psychology frameworks \cite{Costa1999}, and integrated both cognitive and noncognitive aspects into LLM-based personality-aware student simulation \cite{liu-etal-2024-personality}. This augmentation enabled the dialogue model to dynamically adjust its pedagogical approach, providing encouragement for students exhibiting low confidence or being distracted from the on-going session, as shown in Figure \ref{fig:elo_dialogue}.

Moreover, we observed that the scaffolding-guided training improves the dialogue model's robustness regarding inappropriate language and random user inputs. By incorporating dialogic teaching principles and personality-aware student simulation, our system maintains focus on educational objectives and avoids generating harmful or off-topic responses. For instance, when faced with unexpected user behaviors, the model usually prompts the students back to the image description task (i.e., adopting the scaffolding type ``instruction'').

\subsection{Improving Multilingual TTS}
For engaging speech generation, we utilized VITS \cite{kim2021conditional}, a non-autoregressive framework that achieves a balance between speech quality and computational efficiency.
In particular, for low-resource scenarios Malay and Tamil, we collected recordings from adult teachers and children for modeling appropriate prosodic patterns and speech rhythms. The Malay training data includes 22 hours of adult speech from 1 speaker and 9 hours of child speech from 97 speakers, while the Tamil training data comprises 63 hours of adult speech from 1 speaker and 1.5 hours of child speech from 52 speakers.
Speaker embeddings are in a one-hot input format, followed by embedding layer, enabling multi-speaker generation. This approach addresses the issue of voice naturalness in educational contexts, as our preliminary testing revealed that students engage more effectively with systems that generate age-appropriate speech.

\begin{figure}[t!]
\centering
\includegraphics[width=1.0\linewidth]{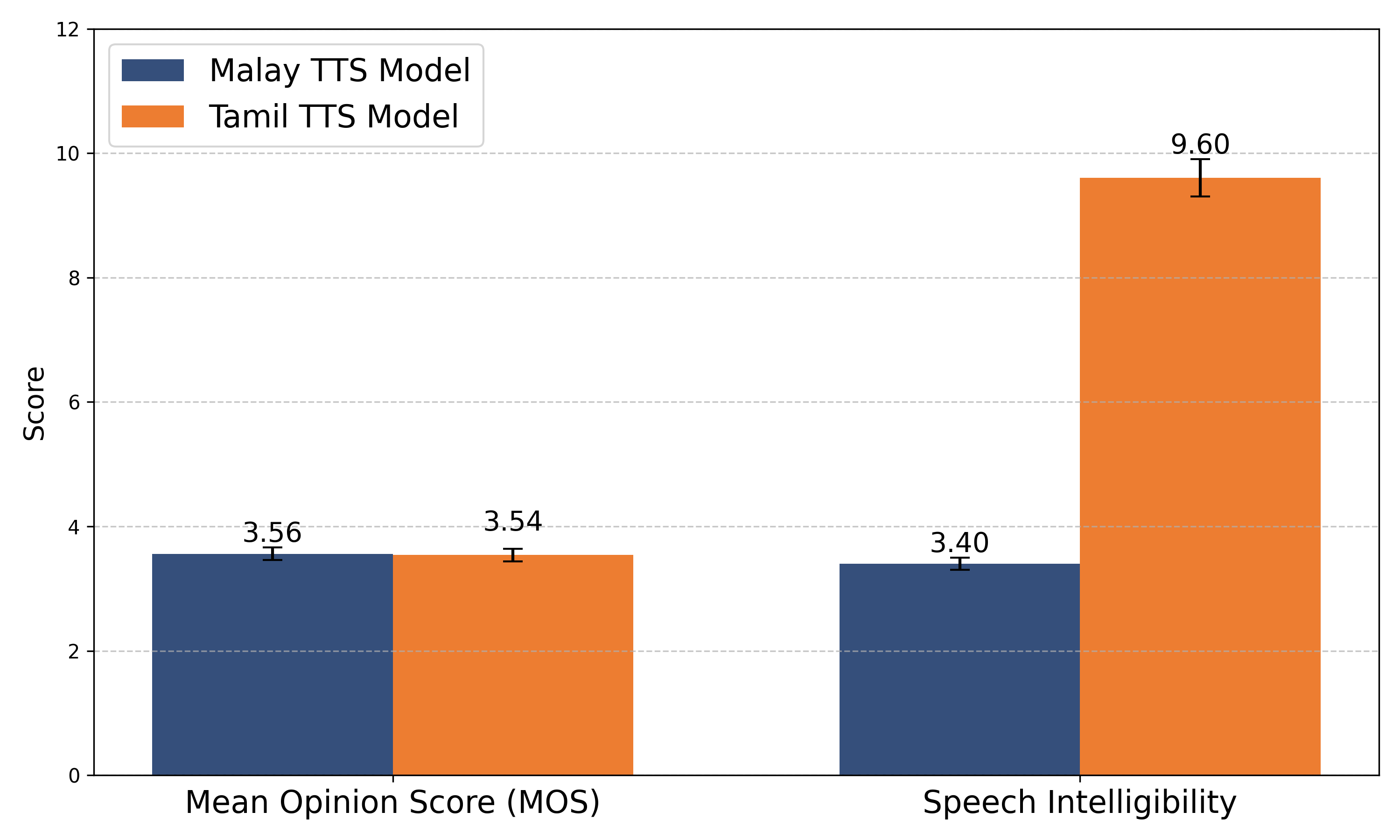}
\caption{Subjective evaluation results for mean opinion score (MOS) and objective evaluation results for speech intelligibility with 95\% confidence intervals for the synthesized Malay and Tamil speech samples by the proposed educational multilingual TTS.}
\label{fig:tts_mos}
\end{figure}

As shown in Figure~\ref{fig:tts_mos}, both objective and subjective evaluations are conducted to assess the multilingual TTS performance. Subjective evaluation is conducted using Mean Opinion Score (MOS) ratings, where listeners assess the overall speech quality and naturalness of the synthesized speech on a 1-5 scoring. It includes 15 Malay and 15 Tamil child speech samples, rated by 20 native Malay and Tamil listeners respectively. The MOS results indicate that our TTS models achieved a reasonable performance (with an average score exceeding 3.50), showing the effectiveness of multilingual adaptation on naturalness and overall quality.
For objective evaluations, we measure speech intelligibility using the widely adopted character error rate (CER). Specifically, we used pretrained Malay and Tamil ASR models to transcribe the Malay and Tamil TTS generation, and computed the CER to quantify speech intelligibility. The results (see Figure \ref{fig:tts_mos}) demonstrate that our TTS models achieve high speech intelligibility, with recognition accuracy exceeding 90\% for both Malay and Tamil.

\begin{figure}[t!]
\centering
\includegraphics[width=1.0\linewidth]{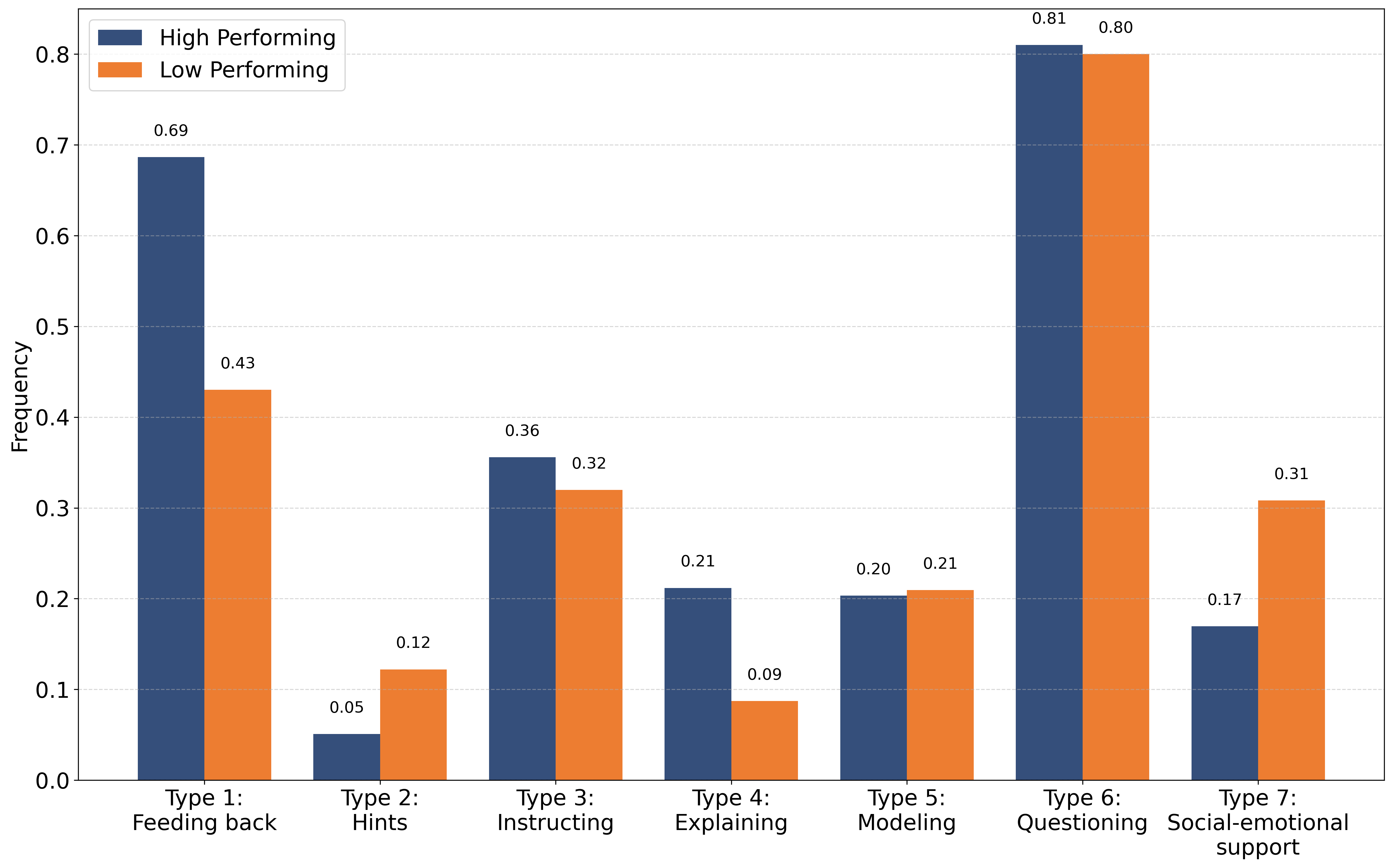}
\caption{Dialogue analysis on scaffolding types of high-performing and low-performing students.}
\label{fig:scaffolding}
\end{figure}

\section{Student Practice and Discussion}
We conducted a user study with 35 elementary school students (grade 1-2) to evaluate SingaKids' effectiveness in real-world educational settings (IRB number: \texttt{IRB-2024-218}). Participants represented diverse language proficiency levels, and they were using the system under the consent and guidance from their parents. Following previous work \cite{liu2024scaffolding}, we conducted a utterance-level analysis of the 7 scaffolding types. As shown in Figure \ref{fig:scaffolding}, significant differences are in some scaffolding types. High-performing students receive more feeding back (69\% vs. 43\%) and explanations (21\% vs. 9\%), where students were encouraged toward deeper understanding. Low-performing students received more hints (12\% vs. 5\%) and social-emotional support (31\% vs. 17\%); the system provides clues, support building confidence when learners struggle.

Moreover, there are observations from our preliminary study: (1) In some cases, noisy environments and children's speech led to more ASR errors, affecting the communication quality. Noise-robust speech recognition and speaker recognition and diarization can help mitigate these issues; (2) Even with dynamic scaffolding and social-emotional support, some students exited sessions when facing persistent difficulties. The scaffolding type ``\textit{Modeling}'' needs to be triggered to prevent frustration; (3) For lower elementary grades, parent guidance is necessary, as they can provide assistance and additional support; (4) When there are many objects and activities in the picture, kids sometimes become distracted or have difficulty pinpointing the focus area, and adding visual highlighting (e.g., bounding boxes) helps improve focus and comprehension. These findings underline the importance of modeling kids-specific learning preferences, to create a more inclusive and effective language learning experience.

\section{Conclusion}
In this work, we presented SingaKids, a multilingual multimodal dialogic tutor designed to enhance elementary language acquisition through picture description tasks. By integrating dense image captioning, multilingual interaction, speech understanding, and engaging speech generation across four languages, our system creates an interactive learning environment that adapts to diverse linguistic contexts. Considering the speech and language proficiency and learning objectives of elementary students, we further improved the system on task-specific optimization and age-appropriate pedagogical alignment. Preliminary empirical studies with elementary school students demonstrated SingaKids' effectiveness in providing self-adaptive guidance through dynamic scaffolding and social-emotional support. Our work provides both technical and educational insights to build general agents in broader educational contexts.

\section*{Limitations}
We are aware that it remains an open problem to mitigate hallucinations and biases in large language models, which may cause communication issues in human-machine interaction and computer-assisted education. Of course, current models and laboratory experiments are always limited in this or similar ways. We do not foresee any unethical uses of our proposed methods or their underlying tools, but hope that it will contribute to reducing incorrect system outputs.

\section*{Ethics and Impact Statement}
We acknowledge that all of the co-authors of this work are aware of the provided ACL Code of Ethics and honor the code of conduct. In our experiments, models are applied under proper license. All data used in this work are only for academic research purposes and should not be used outside of academic research contexts. Our proposed methodology in general does not create a direct societal consequence and are intended to be used to improve the performance, robustness, and safety of the intelligent tutoring systems.

\section*{Acknowledgments}
This research is supported by the AI4EDU Programme in the Institute for Infocomm Research (I$^2$R), Agency for Science, Technology and Research (A*STAR), and the National Research Foundation, Singapore under its AISG Programme (AISG2-GC-2022-005). We thank the anonymous reviewers for their precious feedback to help improve and extend this piece of work.
We acknowledge valuable support and assistance from Siti Umairah Md Salleh, Siti Maryam Binte Ahmad Subaidi, Nabilah Binte Md Johan, Amudha Narayanan, and Anitha Veeramani at the Institute for Infocomm Research (I$^2$R), and valuable contribution in research discussion and study coordination from Chong Han, Audi Arwani Binte Azlan, and Sumi Baby Thomas at the National Institute of Education (NIE), Singapore.

\bibliography{custom}

\begin{thebibliography}{57}
\expandafter\ifx\csname natexlab\endcsname\relax\def\natexlab#1{#1}\fi

\bibitem[{Achiam et~al.(2023{\natexlab{a}})Achiam, Adler, Agarwal, Ahmad, Akkaya, Aleman, Almeida, Altenschmidt, Altman, Anadkat et~al.}]{achiam2023gpt4}
Josh Achiam, Steven Adler, Sandhini Agarwal, Lama Ahmad, Ilge Akkaya, Florencia~Leoni Aleman, Diogo Almeida, Janko Altenschmidt, Sam Altman, Shyamal Anadkat, et~al. 2023{\natexlab{a}}.
\newblock Gpt-4 technical report.
\newblock \emph{arXiv preprint arXiv:2303.08774}.

\bibitem[{Achiam et~al.(2023{\natexlab{b}})Achiam, Adler, Agarwal, Ahmad, Akkaya, Aleman, Almeida, Altenschmidt, Altman, Anadkat et~al.}]{achiam2023gpt}
Josh Achiam, Steven Adler, Sandhini Agarwal, Lama Ahmad, Ilge Akkaya, Florencia~Leoni Aleman, Diogo Almeida, Janko Altenschmidt, Sam Altman, Shyamal Anadkat, et~al. 2023{\natexlab{b}}.
\newblock Gpt-4 technical report.
\newblock \emph{arXiv preprint arXiv:2303.08774}.

\bibitem[{Afzal et~al.(2019)Afzal, Dhamecha, Mukhi, Sindhgatta, Marvaniya, Ventura, and Yarbro}]{afzal-etal-2019-development}
Shazia Afzal, Tejas Dhamecha, Nirmal Mukhi, Renuka Sindhgatta, Smit Marvaniya, Matthew Ventura, and Jessica Yarbro. 2019.
\newblock \href {https://doi.org/10.18653/v1/N19-2015} {Development and deployment of a large-scale dialog-based intelligent tutoring system}.
\newblock In \emph{Proceedings of the NAACL 2019}, pages 114--121, Minneapolis, Minnesota. Association for Computational Linguistics.

\bibitem[{Alexander(2006)}]{Alexander2006}
Robin Alexander. 2006.
\newblock \href {https://books.google.com.sg/books?id=DRoctAEACAAJ} {\emph{{Education as Dialogue: Moral and Pedagogical Choices for a Runaway World}}}.
\newblock Hong Kong Institute of Education.

\bibitem[{Arnau-Gonz{\'a}lez et~al.(2023)Arnau-Gonz{\'a}lez, Arevalillo-Herr{\'a}ez, Albornoz-De~Luise, and Arnau}]{arnau2023methodological}
Pablo Arnau-Gonz{\'a}lez, Miguel Arevalillo-Herr{\'a}ez, Romina Albornoz-De~Luise, and David Arnau. 2023.
\newblock A methodological approach to enable natural language interaction in an intelligent tutoring system.
\newblock \emph{Computer Speech \& Language}, 81:101516.

\bibitem[{Attia et~al.(2024)Attia, Liu, Ai, Demszky, and Espy-Wilson}]{attia2024kid}
Ahmed~Adel Attia, Jing Liu, Wei Ai, Dorottya Demszky, and Carol Espy-Wilson. 2024.
\newblock Kid-whisper: Towards bridging the performance gap in automatic speech recognition for children vs. adults.
\newblock In \emph{Proceedings of the AAAI/ACM Conference on AI, Ethics, and Society}, volume~7, pages 74--80.

\bibitem[{Bai et~al.(2023)Bai, Bai, Chu, Cui, Dang, Deng, Fan, Ge, Han, Huang, Hui, Ji, Li, Lin, Lin, Liu, Liu, Lu, Lu, Ma, Men, Ren, Ren, Tan, Tan, Tu, Wang, Wang, Wang, Wu, Xu, Xu, Yang, Yang, Yang, Yang, Yao, Yu, Yuan, Yuan, Zhang, Zhang, Zhang, Zhang, Zhou, Zhou, Zhou, and Zhu}]{qwen}
Jinze Bai, Shuai Bai, Yunfei Chu, Zeyu Cui, Kai Dang, Xiaodong Deng, Yang Fan, Wenbin Ge, Yu~Han, Fei Huang, Binyuan Hui, Luo Ji, Mei Li, Junyang Lin, Runji Lin, Dayiheng Liu, Gao Liu, Chengqiang Lu, Keming Lu, Jianxin Ma, Rui Men, Xingzhang Ren, Xuancheng Ren, Chuanqi Tan, Sinan Tan, Jianhong Tu, Peng Wang, Shijie Wang, Wei Wang, Shengguang Wu, Benfeng Xu, Jin Xu, An~Yang, Hao Yang, Jian Yang, Shusheng Yang, Yang Yao, Bowen Yu, Hongyi Yuan, Zheng Yuan, Jianwei Zhang, Xingxuan Zhang, Yichang Zhang, Zhenru Zhang, Chang Zhou, Jingren Zhou, Xiaohuan Zhou, and Tianhang Zhu. 2023.
\newblock Qwen technical report.
\newblock \emph{arXiv preprint arXiv:2309.16609}.

\bibitem[{Bhat et~al.(2023)Bhat, Birkl, Wofk, Wonka, and M{\"u}ller}]{bhat2023zoedepth}
Shariq~Farooq Bhat, Reiner Birkl, Diana Wofk, Peter Wonka, and Matthias M{\"u}ller. 2023.
\newblock Zoedepth: Zero-shot transfer by combining relative and metric depth.
\newblock \emph{arXiv preprint arXiv:2302.12288}.

\bibitem[{Chaffar and Frasson(2004)}]{Chaffar2004}
Soumaya Chaffar and Claude Frasson. 2004.
\newblock Inducing optimal emotional state for learning in intelligent tutoring systems.
\newblock In \emph{International Conference on Intelligent Tutoring Systems}, pages 45--54. Springer.

\bibitem[{Chen et~al.(2024)Chen, Wu, Wang, Su, Chen, Xing, Zhong, Zhang, Zhu, Lu et~al.}]{chen2024internvl}
Zhe Chen, Jiannan Wu, Wenhai Wang, Weijie Su, Guo Chen, Sen Xing, Muyan Zhong, Qinglong Zhang, Xizhou Zhu, Lewei Lu, et~al. 2024.
\newblock Internvl: Scaling up vision foundation models and aligning for generic visual-linguistic tasks.
\newblock In \emph{Proceedings of the IEEE/CVF conference on computer vision and pattern recognition}, pages 24185--24198.

\bibitem[{Costa and McCrae(1999)}]{Costa1999}
Paul~T Costa and Robert~R McCrae. 1999.
\newblock A five-factor theory of personality.
\newblock \emph{The five-factor model of personality: Theoretical perspectives}, 2:51--87.

\bibitem[{de~Oliveira et~al.(2023)de~Oliveira, Jones, and Smith}]{de2023interactional}
Luciana~C de~Oliveira, Loren Jones, and Sharon~L Smith. 2023.
\newblock Interactional scaffolding in a first-grade classroom through the teaching--learning cycle.
\newblock \emph{International Journal of Bilingual Education and Bilingualism}, 26(3):270--288.

\bibitem[{Demszky and Hill(2023)}]{demszky2023ncte}
Dorottya Demszky and Heather Hill. 2023.
\newblock The ncte transcripts: A dataset of elementary math classroom transcripts.
\newblock In \emph{Proceedings of the 18th Workshop on Innovative Use of NLP for Building Educational Applications (BEA 2023)}, pages 528--538.

\bibitem[{Dzikovska et~al.(2014)Dzikovska, Steinhauser, Farrow, Moore, and Campbell}]{Dzikovska2014}
Myroslava Dzikovska, Natalie Steinhauser, Elaine Farrow, Johanna Moore, and Gwendolyn Campbell. 2014.
\newblock \href {https://doi.org/10.1007/s40593-014-0017-9} {{BEETLE II: Deep Natural Language Understanding and Automatic Feedback Generation for Intelligent Tutoring in Basic Electricity and Electronics}}.
\newblock \emph{International Journal of Artificial Intelligence in Education}, 24(3):284--332.

\bibitem[{Genishi and Dyson(2015)}]{genishi2015children}
Celia Genishi and Anne~Haas Dyson. 2015.
\newblock \emph{Children, language, and literacy: Diverse learners in diverse times}.
\newblock Teachers College Press.

\bibitem[{Graesser et~al.(2018)Graesser, Hu, and Sottilare}]{Graesser2018}
Arthur~C Graesser, Xiangen Hu, and Robert Sottilare. 2018.
\newblock Intelligent tutoring systems.
\newblock In \emph{International handbook of the learning sciences}, pages 246--255. Routledge.

\bibitem[{Grawemeyer et~al.(2016)Grawemeyer, Mavrikis, Holmes, Gutierrez-Santos, Wiedmann, and Rummel}]{Grawemeyer2016}
Beate Grawemeyer, Manolis Mavrikis, Wayne Holmes, Sergio Gutierrez-Santos, Michael Wiedmann, and Nikol Rummel. 2016.
\newblock \href {https://doi.org/10.1145/2883851.2883936} {Affecting off-task behaviour: how affect-aware feedback can improve student learning}.
\newblock In \emph{Proceedings of the Sixth International Conference on Learning Analytics \& Knowledge}, LAK '16, page 104–113, New York, NY, USA. Association for Computing Machinery.

\bibitem[{Grivokostopoulou et~al.(2017)Grivokostopoulou, Perikos, and Hatzilygeroudis}]{Grivokostopoulou2017}
Foteini Grivokostopoulou, Isidoros Perikos, and Ioannis Hatzilygeroudis. 2017.
\newblock \href {https://doi.org/10.1007/s40593-016-0116-x} {{An Educational System for Learning Search Algorithms and Automatically Assessing Student Performance}}.
\newblock \emph{International Journal of Artificial Intelligence in Education}, 27(1):207--240.

\bibitem[{Harley et~al.(2015)Harley, Bouchet, Hussain, Azevedo, and Calvo}]{Harley2015}
Jason~M. Harley, François Bouchet, M.~Sazzad Hussain, Roger Azevedo, and Rafael Calvo. 2015.
\newblock \href {https://doi.org/https://doi.org/10.1016/j.chb.2015.02.013} {A multi-componential analysis of emotions during complex learning with an intelligent multi-agent system}.
\newblock \emph{Computers in Human Behavior}, 48:615--625.

\bibitem[{Ji et~al.(2023)Ji, Han, and Ko}]{ji2023systematic}
Hyangeun Ji, Insook Han, and Yujung Ko. 2023.
\newblock A systematic review of conversational ai in language education: Focusing on the collaboration with human teachers.
\newblock \emph{Journal of Research on Technology in Education}, 55(1):48--63.

\bibitem[{Johnson et~al.(2016)Johnson, Karpathy, and Fei-Fei}]{densecap}
Justin Johnson, Andrej Karpathy, and Li~Fei-Fei. 2016.
\newblock Densecap: Fully convolutional localization networks for dense captioning.
\newblock In \emph{Proceedings of the IEEE Conference on Computer Vision and Pattern Recognition}.

\bibitem[{Kim et~al.(2021)Kim, Kong, and Son}]{kim2021conditional}
Jaehyeon Kim, Jungil Kong, and Juhee Son. 2021.
\newblock Conditional variational autoencoder with adversarial learning for end-to-end text-to-speech.
\newblock In \emph{International Conference on Machine Learning}, pages 5530--5540. PMLR.

\bibitem[{Kirillov et~al.(2023)Kirillov, Mintun, Ravi, Mao, Rolland, Gustafson, Xiao, Whitehead, Berg, Lo et~al.}]{kirillov2023segment}
Alexander Kirillov, Eric Mintun, Nikhila Ravi, Hanzi Mao, Chloe Rolland, Laura Gustafson, Tete Xiao, Spencer Whitehead, Alexander~C Berg, Wan-Yen Lo, et~al. 2023.
\newblock Segment anything.
\newblock In \emph{Proceedings of the IEEE/CVF international conference on computer vision}, pages 4015--4026.

\bibitem[{Kulik and Fletcher(2016)}]{Kulik2016}
James~A. Kulik and J.~D. Fletcher. 2016.
\newblock \href {https://doi.org/10.3102/0034654315581420} {{Effectiveness of Intelligent Tutoring Systems: A Meta-Analytic Review}}.
\newblock \emph{Review of Educational Research}, 86(1):42--78.

\bibitem[{Leong et~al.(2022)Leong, Nemecek, Mansdorfer, Filighera, Owodunni, and Whitenack}]{leong2022bloom}
Colin Leong, Joshua Nemecek, Jacob Mansdorfer, Anna Filighera, Abraham Owodunni, and Daniel Whitenack. 2022.
\newblock Bloom library: Multimodal datasets in 300+ languages for a variety of downstream tasks.
\newblock \emph{arXiv preprint arXiv:2210.14712}.

\bibitem[{Lin et~al.(2025)Lin, Wang, Liu, and Chen}]{lin2025crossin}
Geyu Lin, Bin Wang, Zhengyuan Liu, and Nancy Chen. 2025.
\newblock Crossin: An efficient instruction tuning approach for cross-lingual knowledge alignment.
\newblock In \emph{Proceedings of the Second Workshop on Scaling Up Multilingual \& Multi-Cultural Evaluation}, pages 12--23.

\bibitem[{Liu et~al.(2024{\natexlab{a}})Liu, Zeng, Ren, Li, Zhang, Yang, Jiang, Li, Yang, Su et~al.}]{liu2024grounding}
Shilong Liu, Zhaoyang Zeng, Tianhe Ren, Feng Li, Hao Zhang, Jie Yang, Qing Jiang, Chunyuan Li, Jianwei Yang, Hang Su, et~al. 2024{\natexlab{a}}.
\newblock Grounding dino: Marrying dino with grounded pre-training for open-set object detection.
\newblock In \emph{European Conference on Computer Vision}, pages 38--55. Springer.

\bibitem[{Liu et~al.(2022)Liu, Ni, Aw, and Chen}]{liu2022singlish}
Zhengyuan Liu, Shikang Ni, Aiti Aw, and Nancy Chen. 2022.
\newblock Singlish message paraphrasing: A joint task of creole translation and text normalization.
\newblock In \emph{Proceedings of the 29th International Conference on Computational Linguistics}, pages 3924--3936.

\bibitem[{Liu et~al.(2024{\natexlab{b}})Liu, Yin, and Chen}]{liu-etal-2024-optimizing}
Zhengyuan Liu, Stella~Xin Yin, and Nancy Chen. 2024{\natexlab{b}}.
\newblock \href {https://doi.org/10.18653/v1/2024.sigdial-1.43} {Optimizing code-switching in conversational tutoring systems: A pedagogical framework and evaluation}.
\newblock In \emph{Proceedings of the 25th Annual Meeting of the Special Interest Group on Discourse and Dialogue}, pages 500--515, Kyoto, Japan. Association for Computational Linguistics.

\bibitem[{Liu et~al.(2024{\natexlab{c}})Liu, Yin, Lee, and Chen}]{liu2024scaffolding}
Zhengyuan Liu, Stella~Xin Yin, Carolyn Lee, and Nancy~F Chen. 2024{\natexlab{c}}.
\newblock Scaffolding language learning via multi-modal tutoring systems with pedagogical instructions.
\newblock In \emph{2024 IEEE Conference on Artificial Intelligence (CAI)}, pages 1258--1265. IEEE.

\bibitem[{Liu et~al.(2024{\natexlab{d}})Liu, Yin, Lin, and Chen}]{liu-etal-2024-personality}
Zhengyuan Liu, Stella~Xin Yin, Geyu Lin, and Nancy~F. Chen. 2024{\natexlab{d}}.
\newblock \href {https://doi.org/10.18653/v1/2024.emnlp-main.37} {Personality-aware student simulation for conversational intelligent tutoring systems}.
\newblock In \emph{Proceedings of the 2024 Conference on Empirical Methods in Natural Language Processing}, pages 626--642, Miami, Florida, USA. Association for Computational Linguistics.

\bibitem[{Macina et~al.(2023{\natexlab{a}})Macina, Daheim, Chowdhury, Sinha, Kapur, Gurevych, and Sachan}]{macina2023mathdial}
Jakub Macina, Nico Daheim, Sankalan Chowdhury, Tanmay Sinha, Manu Kapur, Iryna Gurevych, and Mrinmaya Sachan. 2023{\natexlab{a}}.
\newblock Mathdial: A dialogue tutoring dataset with rich pedagogical properties grounded in math reasoning problems.
\newblock In \emph{Findings of EMNLP 2023}, pages 5602--5621.

\bibitem[{Macina et~al.(2023{\natexlab{b}})Macina, Daheim, Wang, Sinha, Kapur, Gurevych, and Sachan}]{macina2023opportunities}
Jakub Macina, Nico Daheim, Lingzhi Wang, Tanmay Sinha, Manu Kapur, Iryna Gurevych, and Mrinmaya Sachan. 2023{\natexlab{b}}.
\newblock Opportunities and challenges in neural dialog tutoring.
\newblock In \emph{Proceedings of the 17th Conference of the European Chapter of the Association for Computational Linguistics}, pages 2357--2372.

\bibitem[{MacLellan and Koedinger(2022)}]{maclellan2022domain}
Christopher~J MacLellan and Kenneth~R Koedinger. 2022.
\newblock Domain-general tutor authoring with apprentice learner models.
\newblock \emph{International Journal of Artificial Intelligence in Education}, 32(1):76--117.

\bibitem[{Muennighoff et~al.(2023)Muennighoff, Wang, Sutawika, Roberts, Biderman, Le~Scao, Bari, Shen, Yong, Schoelkopf et~al.}]{muennighoff2023crosslingual}
Niklas Muennighoff, Thomas Wang, Lintang Sutawika, Adam Roberts, Stella Biderman, Teven Le~Scao, M~Saiful Bari, Sheng Shen, Zheng~Xin Yong, Hailey Schoelkopf, et~al. 2023.
\newblock Crosslingual generalization through multitask finetuning.
\newblock In \emph{Proceedings of the 61st Annual Meeting of the Association for Computational Linguistics (Volume 1: Long Papers)}, pages 15991--16111.

\bibitem[{Muennighoff et~al.(2022)Muennighoff, Wang, Sutawika, Roberts, Biderman, Scao, Bari, Shen, Yong, Schoelkopf et~al.}]{muennighoff2022crosslingual}
Niklas Muennighoff, Thomas Wang, Lintang Sutawika, Adam Roberts, Stella Biderman, Teven~Le Scao, M~Saiful Bari, Sheng Shen, Zheng-Xin Yong, Hailey Schoelkopf, et~al. 2022.
\newblock Crosslingual generalization through multitask finetuning.
\newblock \emph{arXiv preprint arXiv:2211.01786}.

\bibitem[{Nihad et~al.(2017)Nihad, El~Mokhtar, and Seghroucheni}]{Nihad_El2017}
Elghouch Nihad, En-naimi El~Mokhtar, and Yassine~Zaoui Seghroucheni. 2017.
\newblock \href {https://doi.org/10.3991/ijet.v12i03.6377} {Analysing the outcome of a learning process conducted within the system als\_corr(lp)}.
\newblock \emph{International Journal of Emerging Technologies in Learning (iJET)}, 12(03):pp. 43--56.

\bibitem[{Nye et~al.(2023)Nye, Mee, and Core}]{nye2023generative}
B~Nye, Dillon Mee, and Mark~G Core. 2023.
\newblock Generative large language models for dialog-based tutoring: An early consideration of opportunities and concerns.
\newblock In \emph{AIED Workshops}.

\bibitem[{Nye et~al.(2014)Nye, Graesser, and Hu}]{nye2014autotutor}
Benjamin~D Nye, Arthur~C Graesser, and Xiangen Hu. 2014.
\newblock Autotutor and family: A review of 17 years of natural language tutoring.
\newblock \emph{International Journal of Artificial Intelligence in Education}, 24:427--469.

\bibitem[{Paladines and Ramirez(2020)}]{paladines2020systematic}
Jos{\'e} Paladines and Jaime Ramirez. 2020.
\newblock A systematic literature review of intelligent tutoring systems with dialogue in natural language.
\newblock \emph{IEEE Access}, 8:164246--164267.

\bibitem[{Penedo et~al.(2024)Penedo, Kydl{\'\i}{\v{c}}ek, allal, Lozhkov, Mitchell, Raffel, Werra, and Wolf}]{penedo2024fineweb}
Guilherme Penedo, Hynek Kydl{\'\i}{\v{c}}ek, Loubna~Ben allal, Anton Lozhkov, Margaret Mitchell, Colin Raffel, Leandro~Von Werra, and Thomas Wolf. 2024.
\newblock \href {https://openreview.net/forum?id=n6SCkn2QaG} {The fineweb datasets: Decanting the web for the finest text data at scale}.
\newblock In \emph{The Thirty-eight Conference on Neural Information Processing Systems Datasets and Benchmarks Track}.

\bibitem[{Pokriv{\v{c}}{\'a}kov{\'a}(2019)}]{pokrivvcakova2019preparing}
Silvia Pokriv{\v{c}}{\'a}kov{\'a}. 2019.
\newblock Preparing teachers for the application of ai-powered technologies in foreign language education.
\newblock \emph{Journal of language and cultural education}.

\bibitem[{Radford et~al.(2022)Radford, Kim, Xu, Brockman, McLeavey, and Sutskever}]{radford2022whisper}
Alec Radford, Jong~Wook Kim, Tao Xu, Greg Brockman, Christine McLeavey, and Ilya Sutskever. 2022.
\newblock \href {https://doi.org/10.48550/ARXIV.2212.04356} {Robust speech recognition via large-scale weak supervision}.

\bibitem[{Ruan et~al.(2019)Ruan, Jiang, Xu, Tham, Qiu, Zhu, Murnane, Brunskill, and Landay}]{ruan2019quizbot}
Sherry Ruan, Liwei Jiang, Justin Xu, Bryce Joe-Kun Tham, Zhengneng Qiu, Yeshuang Zhu, Elizabeth~L Murnane, Emma Brunskill, and James~A Landay. 2019.
\newblock Quizbot: A dialogue-based adaptive learning system for factual knowledge.
\newblock In \emph{Proceedings of the 2019 CHI conference on human factors in computing systems}, pages 1--13.

\bibitem[{Sonkar et~al.(2023)Sonkar, Liu, Mallick, and Baraniuk}]{sonkar2023class}
Shashank Sonkar, Naiming Liu, Debshila Mallick, and Richard Baraniuk. 2023.
\newblock Class: A design framework for building intelligent tutoring systems based on learning science principles.
\newblock In \emph{Findings of EMNLP 2023}, pages 1941--1961.

\bibitem[{Stasaski et~al.(2020)Stasaski, Kao, and Hearst}]{stasaski2020cima}
Katherine Stasaski, Kimberly Kao, and Marti~A Hearst. 2020.
\newblock Cima: A large open access dialogue dataset for tutoring.
\newblock In \emph{Proceedings of the Fifteenth Workshop on Innovative Use of NLP for Building Educational Applications}, pages 52--64.

\bibitem[{Sun et~al.(2024)Sun, Liu, Zhou, Huang, Song, Zhao, Zhang, Zhang, and Gai}]{sun2024parrot}
Yuchong Sun, Che Liu, Kun Zhou, Jinwen Huang, Ruihua Song, Wayne~Xin Zhao, Fuzheng Zhang, Di~Zhang, and Kun Gai. 2024.
\newblock Parrot: Enhancing multi-turn instruction following for large language models.
\newblock In \emph{Proceedings of the 62nd Annual Meeting of the Association for Computational Linguistics (Volume 1: Long Papers)}, pages 9729--9750.

\bibitem[{Team et~al.(2023)Team, Anil, Borgeaud, Wu, Alayrac, Yu, Soricut, Schalkwyk, Dai, Hauth et~al.}]{team2023gemini}
Gemini Team, Rohan Anil, Sebastian Borgeaud, Yonghui Wu, Jean-Baptiste Alayrac, Jiahui Yu, Radu Soricut, Johan Schalkwyk, Andrew~M Dai, Anja Hauth, et~al. 2023.
\newblock Gemini: a family of highly capable multimodal models.
\newblock \emph{arXiv preprint arXiv:2312.11805}.

\bibitem[{Teknium(2023)}]{OpenHermes-2.5}
Teknium. 2023.
\newblock \href {https://huggingface.co/datasets/teknium/OpenHermes-2.5} {Openhermes 2.5: An open dataset of synthetic data for generalist llm assistants}.

\bibitem[{Wang et~al.(2023{\natexlab{a}})Wang, Liu, Huang, Jiao, Ding, Aw, and Chen}]{wang2023seaeval}
Bin Wang, Zhengyuan Liu, Xin Huang, Fangkai Jiao, Yang Ding, Ai~Ti Aw, and Nancy~F Chen. 2023{\natexlab{a}}.
\newblock Seaeval for multilingual foundation models: From cross-lingual alignment to cultural reasoning.
\newblock \emph{arXiv preprint arXiv:2309.04766}.

\bibitem[{Wang et~al.(2023{\natexlab{b}})Wang, Zhang, Robinson, Loeb, and Demszky}]{wang2023step}
Rose~E Wang, Qingyang Zhang, Carly Robinson, Susanna Loeb, and Dorottya Demszky. 2023{\natexlab{b}}.
\newblock Step-by-step remediation of students' mathematical mistakes.
\newblock \emph{arXiv preprint arXiv:2310.10648}.

\bibitem[{Wong et~al.(2022)Wong, Zhang, and Chen}]{wong2022variations}
Jeremy Heng~Meng Wong, Huayun Zhang, and Nancy~F Chen. 2022.
\newblock Variations of multi-task learning for spoken language assessment.
\newblock In \emph{Interspeech}, pages 4456--4460.

\bibitem[{Xu et~al.(2019)Xu, Wijekumar, Ramirez, Hu, and Irey}]{Xu2019}
Zhihong Xu, Kausalai Wijekumar, Gilbert Ramirez, Xueyan Hu, and Robin Irey. 2019.
\newblock \href {https://doi.org/10.1111/bjet.12758} {{The effectiveness of intelligent tutoring systems on K-12 students' reading comprehension: A meta-analysis}}.
\newblock \emph{British Journal of Educational Technology}, 50(6):3119--3137.

\bibitem[{Yan et~al.(2024)Yan, Sha, Zhao, Li, Martinez-Maldonado, Chen, Li, Jin, and Gašević}]{yan2024practical}
L.~Yan, L.~Sha, L.~Zhao, Y.~Li, R.~Martinez-Maldonado, G.~Chen, X.~Li, Y.~Jin, and D.~Gašević. 2024.
\newblock \href {https://doi.org/10.1111/bjet.13370} {Practical and ethical challenges of large language models in education: A systematic scoping review}.
\newblock \emph{British Journal of Educational Technology}, 55(1):90--112.

\bibitem[{Zhai et~al.(2021)Zhai, Chu, Chai, Jong, Istenic, Spector, Liu, Yuan, and Li}]{zhai2021review}
Xuesong Zhai, Xiaoyan Chu, Ching~Sing Chai, Morris Siu~Yung Jong, Andreja Istenic, Michael Spector, Jia-Bao Liu, Jing Yuan, and Yan Li. 2021.
\newblock A review of artificial intelligence (ai) in education from 2010 to 2020.
\newblock \emph{Complexity}, 2021(1):8812542.

\bibitem[{Zhang et~al.(2021)Zhang, Shi, and Chen}]{zhang2021multilingual}
Huayun Zhang, Ke~Shi, and Nancy~F Chen. 2021.
\newblock Multilingual speech evaluation: Case studies on english, malay and tamil.
\newblock In \emph{Proc. Interspeech 2021}, pages 4443--4447.

\bibitem[{Zhang and Aslan(2021)}]{zhang2021ai}
Ke~Zhang and Ayse~Begum Aslan. 2021.
\newblock Ai technologies for education: Recent research \& future directions.
\newblock \emph{Computers and education: Artificial intelligence}, 2:100025.

\end{thebibliography}




\end{document}